\documentclass[10pt, conference]{IEEEtran}
\usepackage{booktabs}
\usepackage{tabularx}
\usepackage{amsmath}
\usepackage{graphicx}
\usepackage{cite}
\usepackage{orcidlink}
\usepackage{xurl}
\hypersetup{hidelinks}

\begin{document}

\title{Event-Native Symbolic-Temporal Spike Encoding Framework for Heterogeneous Cyber Streams}

\author{
\IEEEauthorblockN{
Dalton Diez\textsuperscript{\orcidlink{0009-0009-6404-1577}},
Peyton Andras\textsuperscript{\orcidlink{0009-0004-5113-9704}},
Max Shroyer\textsuperscript{\orcidlink{0009-0008-8023-6246}}, and
James Ghawaly, Jr.\textsuperscript{\orcidlink{0000-0001-8826-8500}}
}
\IEEEauthorblockA{
Louisiana State University\\
Baton Rouge, Louisiana, USA\\
\{ddiez2, peyton.andras, mshroy1, jghawaly\}@lsu.edu
}
}

\maketitle

\begin{abstract}
Spiking neural networks (SNNs) have shown promise for sparse, event-driven computation through stateful processing that is naturally compatible with low-power edge hardware. These properties align with cyber monitoring, where data arrives asynchronously, and malicious behavior often emerges through temporal patterns across event sequences. However, cyber streams are not composed solely of continuous numeric signals: their informative structure is also carried by categorical identifiers, irregular timing, and local behavioral context. Traditional rate- and population-based spike encodings are not naturally suited to these heterogeneous semantics, while conventional intrusion detection system (IDS) pipelines typically resolve the mismatch by converting raw events into flows, fixed aggregation windows, or dense tensors. Although useful for conventional classifiers, these transformations introduce buffering latency, obscure native temporal structure, and weaken the computational advantages of event-driven neuromorphic processing.

We introduce an event-native symbolic-temporal spike encoding framework that maps heterogeneous cyber events directly into sparse, spike-compatible inputs. By assigning explicit encoding roles to semantic identity, local frequency context, and inter-event timing, the framework preserves categorical semantics and temporal dynamics. We validate the approach on packet-level Network IDS and extend it to message-level CAN IDS, using both domains to evaluate whether the encoding exposes usable structure for recurrent SNNs operating directly on native event streams. Under edge-oriented, $\mu$Caspian-aligned hardware constraints, compact recurrent SNNs achieve strong anomaly detection performance, with an operational hybrid metric ($J_{hybrid}$) of 0.987 on Network IDS and 0.980 on CAN IDS. These results show that preserving native symbolic and temporal structure enables compact recurrent SNNs to perform effective attack detection using edge-compatible inference.

\end{abstract}


\section{Introduction}

Spiking neural networks (SNNs) are naturally aligned with the asynchronous, event-driven characteristics of cybersecurity streams~\cite{maass1997networks}. In domains like network monitoring or in-vehicle communication, data arrives as discrete events (packets or messages), and malicious activity often manifests as temporal patterns across sequences. However, despite this conceptual alignment, applying SNNs to native cyber event streams presents a major representational challenge.

The difficulty lies in the input interface. Unlike physical sensor signals, which are often regularly sampled and continuous-valued, cyber event fields are symbolic, heterogeneous, and irregularly timed. Attributes such as protocol type, destination port, service, and Controller Area Network (CAN) arbitration ID do not have natural numeric ordering or magnitude relationships. Treating a port number or protocol identifier as a continuous value therefore distorts its semantic meaning. Individual packets or messages are also weakly informative in isolation; their significance depends on local sequential context and relative timing across event arrivals.

To overcome this representation gap, conventional intrusion detection systems (IDS) rely on extensive preprocessing, such as flow construction, fixed-duration aggregation, or input aggregation windows~\cite{lakhina2005mining, sommer2010outside, rajapaksha2023cansurvey}. While these methods extract useful numeric features, they transform the data from its native form, introducing latency and increasing resource overhead. Prior SNN-based IDS research typically inherits these limitations by using pre-engineered, static numeric feature vectors as the starting point for spike encoding~\cite{zhou2021spiking, SimicSNN, prajwalasimha, 10.1145/3183584.3183617}, or by converting conventional deep neural networks into spiking equivalents via artificial-to-SNN pipelines~\cite{islam2024unsupervised}. This prevents end-to-end event-native inference and shifts computation back toward preprocessing stages that reduce the latency and low-power advantages of sparse neuromorphic processing.

In this work, we propose an event-to-spike representation designed specifically for heterogeneous, asynchronous cyber streams. Our central claim is representational: cyber event streams can be made suitable for edge-based neuromorphic inference when each event is encoded according to its symbolic identity, local behavioral context, and inter-event timing. Preserving this structure allows compact recurrent SNNs to accumulate evidence over time and classify attacks directly from native cyber streams. The contribution is not a new IDS model, but a symbolic-temporal representation interface for mapping raw cyber events into sparse, spike-compatible channels at their exact moment of arrival. The encoder uses semantically defined bins for categorical attributes, online frequency counters for local context, and discretized inter-event intervals ($\Delta t$) for relative timing. This enables direct edge-compatible event-native classification without flow construction, fixed-duration aggregation, or dense traffic tensors.

To demonstrate the generality of this representation, we instantiate it across two distinct domains: packet-level Network IDS and message-level CAN IDS. We optimize the recurrent SNN topologies and parameters under edge-oriented $\mu$Caspian hardware constraints~\cite{mitchell2020ucaspian} using Evolutionary Optimization for Neuromorphic Systems (EONS)~\cite{schuman2020evolutionary} as the optimization backend. We evaluate the representation with lightweight baselines on raw and encoded inputs, contextualizing the difficulty of event-native detection and the value of the domain-informed encoded representation. We then show that SNNs can use this representation to retain context across native event arrivals and perform effective edge-compatible detection.


\section{Related Work}

\subsection{SNN Input Encoding}
Input encoding converts real-world data into spikes. Traditional encoding schemes (e.g., rate, temporal, and population coding) were designed for regularly sampled, continuous-valued signals where information is carried by physical magnitudes or precise timing~\cite{gerstner2002spiking, bouvier2019spiking}. These approaches are poorly suited for cyber streams, where information is carried by symbolic attributes, irregular event spacing, and local temporal context rather than continuous magnitudes. For this reason, a direct rate-coded or population-coded baseline is not semantically well-defined for native packet or CAN fields. Ports, protocols, services, and arbitration IDs do not encode information through numeric magnitude, so mapping them directly to firing rates or population centers would impose artificial relationships that are not present in the underlying protocol semantics.

\subsection{Event Streams and Irregular Temporal Models}
Outside neuromorphic computing, sequence models like DeepLog capture log sequences~\cite{du2017deeplog}, while temporal point processes model irregular event arrivals~\cite{mei2017neuralhawkes}. Event-based vision pipelines also process asynchronous pixel streams using specialized spatial-temporal representations~\cite{gallego2019eventbasedvision}. While these approaches capture temporal relationships, they rely on dense embeddings, self-attention, or large hidden states that are computationally expensive and incompatible with edge neuromorphic hardware.

\subsection{Machine Learning for Network IDS and CAN IDS}
Conventional Network IDS and CAN IDS research heavily favors static, aggregated, or fixed-dimensional feature representations. Evaluations on network datasets typically use processed traffic features that summarize multiple packets over time~\cite{asgharzadeh2024intrusion,moustafa2021new,SinhaLSTM}. Similarly, CAN IDS benchmarks frequently use fixed-dimensional representations of message-ID sequences or payload bit strings processed by dense neural networks~\cite{song2020carhacking,verma2024road}. These approaches require buffering and aggregation, adding latency and preprocessing overhead that reduces their suitability for direct edge-based event processing.

\subsection{Spiking Neural Networks for Intrusion Detection}
Prior spiking IDS implementations generally inherit these aggregation assumptions. Most spiking network intrusion detectors perform rate or temporal encoding on pre-extracted, high-level numeric features rather than raw packets~\cite{zhou2021spiking, SimicSNN, prajwalasimha}. For CAN security, researchers have proposed ANN-to-SNN conversion methods that map pre-trained networks into spiking networks~\cite{islam2024unsupervised}. Because these networks operate on fixed numeric matrices, they do not process asynchronous cyber events directly and therefore cannot support event-native execution.

\subsection{Representation Gap}
Despite domain alignments, there is a representational gap between asynchronous cyber streams and neuromorphic computation. Current SNN encoders cannot naturally represent the mixture of categorical symbols, irregular timings, and sequential contexts of packet and CAN traffic. Conversely, existing IDS pipelines recover context only after buffering or aggregation, discarding the timing and event boundaries that recurrent SNNs are designed to process. This motivates a representation layer that preserves cyber event semantics while enabling sparse recurrent SNN execution for direct, low-overhead edge-based monitoring.


\section{Symbolic-Temporal Encoding Framework}
\label{sec:encoding}

The core contribution of this work is a representation framework for asynchronous cyber streams that maps raw events into sparse, spike-compatible inputs at their moment of arrival. The encoder represents categorical cyber fields using operational semantics, maintains local context with online frequency counters, and captures timing explicitly, rather than implicitly through fixed-rate intervals.

\subsection{General Framework}
Given an asynchronous event stream $\mathcal{X} = \{(x_1, t_1), (x_2, t_2), \ldots, (x_T, t_T)\}$, each event $x_i$ arriving at timestamp $t_i$ is mapped to a sparse binary spike vector $s_i \in \{0,1\}^D$ by concatenating $K$ attribute groups with one temporal group:
\begin{equation}
\label{eq:token}
  s_i = \bigl[\,g_1(x_i) \;\|\; g_2(x_i) \;\|\; \cdots \;\|\; g_K(x_i) \;\|\; h(\Delta t_i)\,\bigr] \in \{0,1\}^D,
\end{equation}
where $\Delta t_i = t_i - t_{i-1}$ is the inter-event arrival delay.

\begin{figure}[t]
    \centering
    \includegraphics[width=.8\columnwidth]{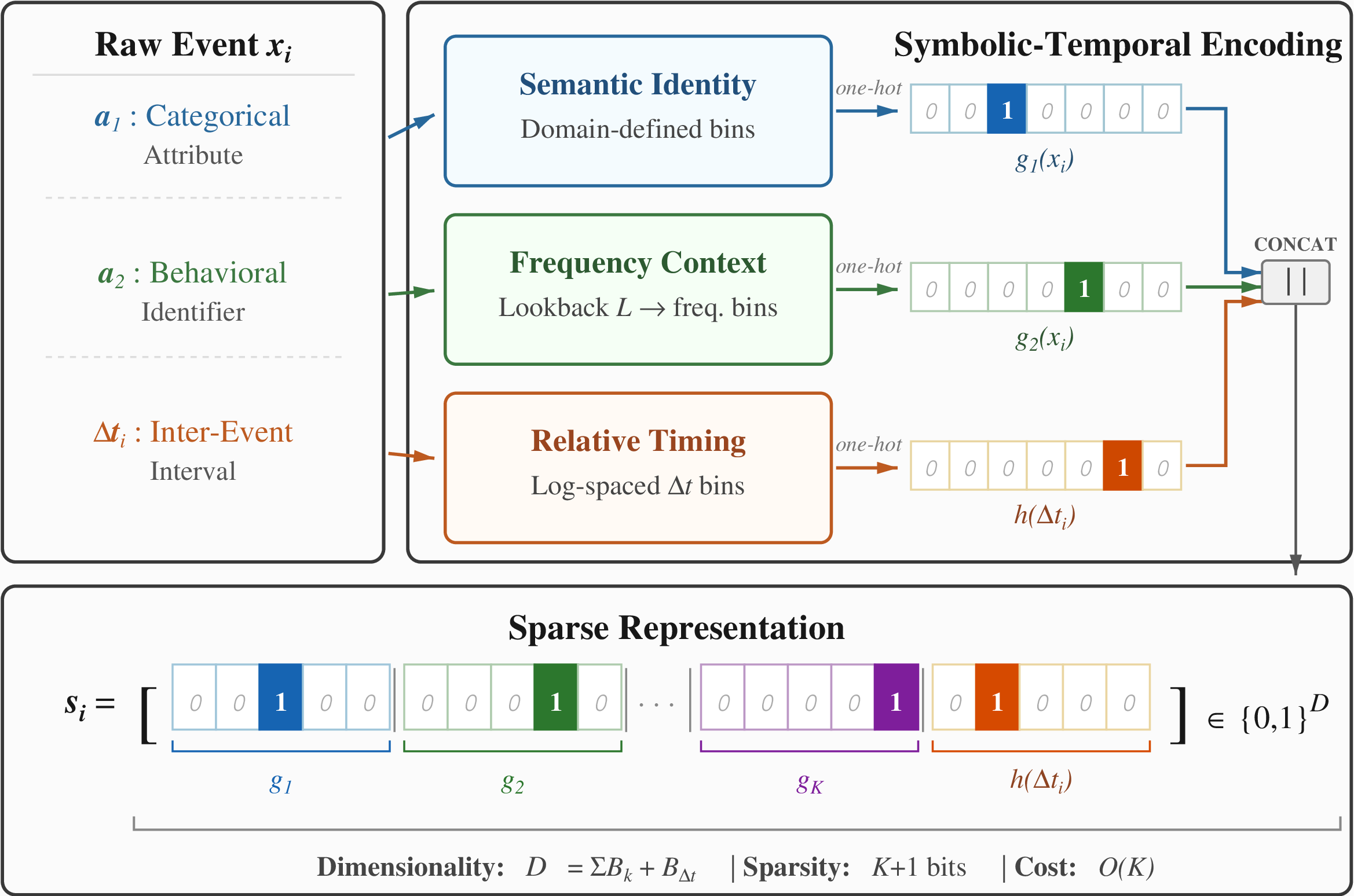}
    \caption{Overview of the symbolic-temporal encoding pipeline, mapping a raw cyber event into a sparse, spike-compatible binary vector.}
    \label{fig:examplevector}
\end{figure}

Each attribute group $g_k(x_i)$ maps a specific categorical or numeric field to a one-hot subvector of size $B_k$. The temporal group $h(\Delta t_i)$ maps the inter-event delay to a one-hot subvector of size $B_{\Delta t}$. The total dimensionality of the input space is:
\begin{equation}
D = \sum_{k=1}^K B_k + B_{\Delta t}.
\end{equation}
Because each group activates exactly one bin per event, the resulting vector $s_i$ contains exactly $K{+}1$ active elements, with the remaining $D - (K{+}1)$ elements set to zero. The encoding operation has a time complexity of $O(K)$ per event, which is independent of the stream length. This mapping is illustrated in Fig.~\ref{fig:examplevector}. This fully categorical formulation is used for Network IDS; Section VI extends it with rate-coded payload channels.

\subsection{Three Encoding Roles}
To capture the varied semantics of cyber protocols, the feature groups are designed to fulfill three distinct representational roles:
\begin{enumerate}
    \item \textbf{Semantic Identity}: Non-numeric categorical fields (such as protocols, services, and ports) are mapped into semantically defined bins based on domain context. This avoids assigning arbitrary continuous magnitudes to categorical values, preserving their symbolic meaning.
    \item \textbf{Frequency-Based Context}: High-cardinality identifiers, such as IP addresses or CAN arbitration IDs, are mapped to local frequency bins. The encoder maintains an online FIFO window of the most recent $L$ events and updates identifier counts as events arrive. The resulting count for the current identifier is discretized into a frequency bin, capturing local burst or periodic behavior without overfitting to specific identifier values.
    \item \textbf{Relative Timing:} The inter-event delay $\Delta t_i$ is mapped to logarithmically spaced bins, allowing the recurrent SNN to learn burst, periodic, and idle-time dynamics directly from relative event spacing without fixed-duration accumulation windows or clock-aligned feature bins.
\end{enumerate}

\subsection{Key Design Properties}
The encoding framework satisfies four properties:
\begin{itemize}
    \item \textbf{Event-Native}: Events are processed individually upon arrival, avoiding flow reconstruction or dense traffic-summary tensors.
    \item \textbf{Sparse}: The resulting representation is highly sparse, activating only $K+1$ channels out of $D$.
    \item \textbf{Domain-Informed}: Bin boundaries are derived from standard network and communication specifications (e.g., IANA port allocations or Ethernet frame sizes).
    \item \textbf{Hardware-Compatible}: The binary activations align directly with the input channels of neuromorphic hardware, minimizing encoding overhead.
\end{itemize}

\section{Experimental Setup}
\label{sec:experimental_setup}

All experiments use a common classification setting, training protocol, hardware-constrained simulator, and evaluation methodology. Data are partitioned at the run level: packets or messages from the same temporal run are never split across training, validation, and test sets. For the Combined datasets, stratification is performed by attack category while preserving run boundaries. This avoids event-level leakage across partitions and ensures that test performance is measured on held-out runs. Training runs are used for evolutionary fitness evaluation. Validation runs are monitored during evolution as a generalization check. Held-out test runs are used only for final performance reporting.

\subsection{Binary Classification Task}
We formulate cyber intrusion detection as a per-event binary classification task. Given an event stream $\mathcal{X}$, the model must output a prediction $\hat{y}_i \in \{0, 1\}$ upon the arrival of each event $x_i$, where $1$ indicates malicious and $0$ indicates benign activity. The model must perform this classification using the current encoded event $s_i$ and the context accumulated within the network's recurrent state.

\subsection{SNN Optimization and Hardware-Constrained Simulation}

We optimize recurrent SNN topology and parameters using Evolutionary Optimization for Neuromorphic Systems (EONS)~\cite{schuman2020evolutionary}. EONS is a population-based, gradient-free optimizer that evolves both network structure and neuron/synapse parameters using tournament selection, crossover, and mutation. EONS serves as the optimization backend; the contribution of this work is the event-native representation supplied to the recurrent SNN. For each dataset, we perform 200 independent EONS runs, each using 300 generations and a population size of 100 candidate networks, for a total of 6.0 million candidate evaluations per dataset.

All candidate networks are evaluated in the Caspian simulator configured to emulate the $\mu$Caspian neuromorphic processor~\cite{mitchell2020ucaspian}. Networks use Integrate-and-Fire neurons with 8-bit thresholds, 16-bit membrane potentials, and 4-bit axonal delays. To maintain edge relevance, evolution is constrained to networks with at most 256 neurons and 4,096 synapses.

\subsection{Rolling Decision Rule}
To prevent transient spikes in output activity from causing excessive false alarms, we apply a rolling decision rule to the raw per-event predictions $\hat{y}_i$. An attack alarm is raised if the number of malicious predictions within a rolling output alarm window of the last $w$ events meets or exceeds a threshold $\theta$:
\begin{equation}
\label{eq:rolling}
  \hat{y}^{{alarm}}_t =
  \begin{cases}
    1 & \text{if } \displaystyle\sum_{i=t-w+1}^{t} \hat{y}_i \geq \theta, \\[6pt]
    0 & \text{otherwise}.
  \end{cases}
\end{equation}
We fix $w = 50$ events across all experiments to suppress isolated false positives while maintaining low detection latency. The threshold $\theta$ is optimized on the training split. This rolling rule does not aggregate input features, construct flows, or build dense traffic-summary tensors.

\subsection{Baseline Models}
For representation analysis, we train Logistic Regression (LR), Random Forest (RF), Decision Tree (DT), Gaussian Naive Bayes (GNB), and Recurrent Neural Network (RNN) classifiers. All baselines use the same train/validation/test partitions, threshold-selection procedure, and output rolling decision rule as the SNN. Raw features are represented using normalized continuous values, while encoded features use the sparse binary input vector.

\subsection{Evaluation Metrics}

Cybersecurity evaluation must account for both whether attacks are detected and how frequently benign traffic produces false alarms. Prior NIDS work provides precedent for evaluating successful detection at the attack-instance level while measuring false positives at the packet level~\cite{bolzoni}. We follow this general distinction for event-stream intrusion detection. We define the Attack Detection Rate (ADR) as the proportion of labeled attack intervals containing at least one alarm, and $FPR_{event}$ as the fraction of benign event positions for which the detector is in an alarm state. For Network IDS, an event is a packet; for CAN IDS, it is a CAN message. Because EONS requires a single fitness objective, we combine them as $J_{hybrid}$.

\begin{equation}
{ADR} =
\frac{N_{{detected\ attacks}}}
{N_{{attacks}}}.
\end{equation}

\begin{equation}
FPR_{event} =
\frac{N_{{benign\ events\ in\ alarm}}}
{N_{{benign\ events}}}.
\end{equation}

\begin{equation}
J_{hybrid} =
{ADR} - FPR_{event}.
\end{equation}
This metric is not conventional Youden's $J$, because ADR and $FPR_{event}$ are measured at different resolutions. Instead, $J_{hybrid}$ is an operational IDS objective that rewards detection of attack instances while penalizing alarm activity on benign traffic. This provides a single fitness value for EONS that balances attack detection and false-positive behavior. We additionally report per-packet or per-message accuracy by comparing the detector alarm state with the ground-truth label at each event position.

\subsection{Confidence Intervals}

We estimate 95\% confidence intervals using 10,000 bootstrap samples of the test set. Clusters are grouped by source dataset and traffic class, then sampled with replacement within each group. For each sample, ADR and $FPR_{event}$ are recomputed from the resampled attack detections and benign event positions, and $J_{hybrid}$ is then recalculated. The confidence interval is given by the 2.5th and 97.5th percentiles of its bootstrap values. The same samples are used for all models to ensure paired comparisons.

\section{Network IDS Evaluation}
\label{sec:network_ids_evaluation}

We first evaluate the symbolic-temporal framework on packet-level network intrusion detection, where malicious behaviors must be identified from sequential packet arrivals.

\subsection{Dataset and Preprocessing}
We utilize three public packet-level network datasets: TON-IoT~\cite{moustafa2021new}, Bot-IoT~\cite{BotIOT}, and TII-SSRC-23~\cite{herzalla2023tii}, and construct a Combined dataset containing runs from all three sources (Table~\ref{tab:net-datasets}). Raw packet captures are parsed using PyShark to extract packet-level attributes and ordered into temporal runs. The dataset is balanced at the run level (equal numbers of benign and attack-containing runs), though individual event labels remain highly imbalanced to reflect realistic deployment. The runs are split 60/10/30 into train, validation, and test partitions, stratified by attack category.

Packet-level IDS is challenging because individual packets are often weakly informative in isolation, and malicious behavior is usually expressed through traffic patterns, feature distributions, and temporal context rather than a single separable packet attribute~\cite{lakhina2005mining,sommer2010outside}. This challenge is amplified by the irregular timing structure of network traffic~\cite{paxson1997wide} and the heterogeneity and imbalance of real IDS datasets~\cite{moustafa2021new,BotIOT,herzalla2023tii}. These properties motivate an event-native representation that preserves per-packet semantics while allowing the recurrent SNN to accumulate evidence across arrivals.

\begin{table}[htbp]
\caption{Network IDS Dataset Summary}
\label{tab:net-datasets}
\centering
\small
\begin{tabular}{lrrr}
\toprule
Dataset & Runs & Packets & Attack \% \\
\midrule
TON-IoT~\cite{moustafa2021new} & 1,346 & 57.4 M & 5.26 \\
Bot-IoT~\cite{BotIOT} & 31,096 & 459.4 M & 12.60 \\
TII-SSRC-23~\cite{herzalla2023tii} & 521 & 2.84 M & 32.48 \\
\midrule
Combined & 32,963 & 519.6 M & 11.90 \\
\bottomrule
\end{tabular}
\end{table}

\subsection{Encoding Scheme}
We map each packet to a 36-dimensional sparse spike vector composed of eight feature groups (Table~\ref{tab:bin_definitions}). IP frequency groups keep a local count of source and destination addresses over a rolling 20,000-packet window, capturing burst behavior without global address tracking. Port groups are binned according to IANA specifications~\cite{rfc6335}, with destination ports having finer granularity to represent common service endpoints. Packet sizes are binned using Ethernet standard frames~\cite{ieee8023}. Inter-event times ($\Delta t$) are discretized into log-spaced bins to capture the heavy-tailed timing distributions typical of network traffic~\cite{paxson1997wide}. 

\begin{table*}[t]
\centering
\small
\caption{Symbolic Input Bin Definitions. Each network packet activates exactly one bin per feature group.}
\label{tab:bin_definitions}
\setlength{\tabcolsep}{10pt}
\renewcommand{\arraystretch}{1.15}
\begin{tabular}{cl|cl|cl}
\toprule
\textbf{Bin} & \textbf{Definition} &
\textbf{Bin} & \textbf{Definition} &
\textbf{Bin} & \textbf{Definition} \\
\midrule
0  & Src IP freq: Low (0--5k)      & 
12 & Dst port: Web (80, 443, 8080) & 
24 & Service: HTTP/HTTPS \\
1  & Src IP freq: Med-low (5k--10k)&
13 & Dst port: DNS (53, 5353)      &
25 & Service: DNS \\
2  & Src IP freq: Med-high (10k--15k)&
14 & Dst port: SSH (22)            &
26 & Service: TLS/SSL \\
3  & Src IP freq: High (15k--20k)  &
15 & Dst port: Mail (25, 465, 587) &
27 & Service: Other \\
4  & Dst IP freq: Low (0--5k)      &
16 & Dst port: NTP (123)           &
28 & Pkt size: Tiny (0--64 B) \\
5  & Dst IP freq: Med-low (5k--10k)&
17 & Dst port: Other well-known    &
29 & Pkt size: Small (65--512 B) \\
6  & Dst IP freq: Med-high (10k--15k)&
18 & Dst port: Registered          &
30 & Pkt size: Standard (513--1514 B) \\
7  & Dst IP freq: High (15k--20k)  &
19 & Dst port: Dynamic/Other       &
31 & Pkt size: Large ($>$1514 B) \\
8  & Src port: Well-known (0--1023)&
20 & Protocol: TCP                 &
32 & $\Delta t$: Very fast ($<$0.1 ms) \\
9  & Src port: Registered          &
21 & Protocol: UDP                 &
33 & $\Delta t$: Fast (0.1--1 ms) \\
10 & Src port: Dynamic             &
22 & Protocol: ICMP                &
34 & $\Delta t$: Moderate (1--10 ms) \\
11 & Src port: Other               &
23 & Protocol: Other               &
35 & $\Delta t$: Slow ($\geq$10 ms) \\
\bottomrule
\end{tabular}
\end{table*}

\subsection{Results and Baseline Comparison}

The Network IDS evaluation assesses whether the symbolic-temporal encoding exposes useful structure from packet-level event streams and whether recurrent spiking state provides additional temporal decision capacity. We therefore report the best evolved SNN performance across datasets and compare raw and encoded inputs using controlled baselines on the Combined Network IDS dataset.

Table~\ref{tab:net-results} reports the best evolved SNN on each Network IDS dataset. For each dataset, the reported network was selected as the highest-performing candidate on the training data and then evaluated on held-out test runs. Performance remains strong across all individual datasets and the Combined setting. The Combined result is particularly important because it merges traffic from multiple sources and attack distributions, making it a stronger test of whether the representation can support detection across heterogeneous packet streams.

\begin{table}[htbp]
\caption{Best SNN Performance per Network IDS Dataset}
\label{tab:net-results}
\centering
\small
\setlength{\tabcolsep}{3pt}
\begin{tabular}{@{}lcccc@{}}
\toprule
Dataset & $J_{hybrid}$ & ADR &
$FPR_{event}$ & Packet Accuracy \\
\midrule
TON-IoT     & 0.9929 & 0.9952 & 0.0023 & 0.9826 \\
Bot-IoT     & 0.9908 & 0.9998 & 0.0090 & 0.8665 \\
TII-SSRC-23 & 0.9996 & 1.0000 & 0.0004 & 0.7952 \\
Combined    & 0.9873 & 0.9986 & 0.0113 & 0.8750 \\
\bottomrule
\end{tabular}
\end{table}

These results are not intended as a direct state-of-the-art comparison against flow-based or heavily engineered IDS pipelines, since published systems often use different preprocessing, feature construction, task definitions, and evaluation settings. For context, prior work reports 99.99\% accuracy on TON-IoT using CNNs~\cite{asgharzadeh2024intrusion}, 99.87\% on Bot-IoT using hybrid LSTM-CNNs~\cite{SinhaLSTM}, and 100\% on TII-SSRC-23 using XGBoost~\cite{herzalla2023tii}. The relevance of Table~\ref{tab:net-results} is instead that edge-aligned SNNs can achieve high detection directly from packet arrivals while avoiding flow construction and dense feature engineering.

Table~\ref{tab:net-baselines} isolates the role of the representation. Classical models trained on raw packet features perform poorly to moderately, indicating that the packet-level task is difficult when events are treated as static numeric records. When the same models are trained on the proposed encoded representation, performance improves for most classifiers. This shows that the encoder exposes class-relevant structure that is not as accessible in the raw numeric feature space. However, the encoded representation alone is not sufficient for strong detection with static classifiers. To provide a non-spiking recurrent-state comparison, we additionally evaluate an RNN using the same encoded event sequences. The RNN achieves $J_{hybrid}=0.739$, improving over the static Random Forest and Decision Tree models ($J_{hybrid}=0.709$), while the recurrent SNN reaches $J_{hybrid}=0.987$ with a substantially lower $FPR_{event}$ of $0.011$, compared with $0.140$ for the RNN.

Detection latency is measured as the number of packets between attack onset and the first alarm within the labeled attack interval. On the Combined dataset, the best individual network achieves a median detection latency of 55 packets, with 25\% of attacks detected within 22 packets. During high-rate packet bursts, this corresponds to short response delays, further supporting the real-time motivation for event-driven neuromorphic intrusion detection.

Together, these results indicate that a major obstacle to applying compact SNNs to cyber monitoring is representational: once native event semantics are preserved, sparse recurrent inference becomes a viable fit for asynchronous, edge-oriented cyber streams.

\begin{table*}[htbp]
\caption{Baseline Comparison on the Combined Network IDS Dataset}
\label{tab:net-baselines}
\centering
\small
\begin{tabular}{llccccc}
\toprule
Model & Representation & $J_{hybrid}$ & $J_{hybrid}$ 95\% CI & ADR & $FPR_{event}$ & Packet Accuracy \\
\midrule
LR  & Raw     & 0.4367 & [0.4150, 0.4584] & 0.8729 & 0.4362 & 0.4698 \\
LR  & Encoded & 0.6811 & [0.6673, 0.6948] & 0.9364 & 0.2554 & 0.6444 \\
RF  & Raw     & 0.7004 & [0.6870, 0.7135] & 0.9078 & 0.2074 & 0.5574 \\
RF  & Encoded & 0.7089 & [0.6960, 0.7218] & 0.9607 & 0.2517 & 0.6709 \\
DT  & Raw     & 0.7068 & [0.6934, 0.7200] & 0.8937 & 0.1869 & 0.5428 \\
DT  & Encoded & 0.7088 & [0.6958, 0.7216] & 0.9607 & 0.2519 & 0.6708 \\
GNB & Raw     & 0.1098 & [0.0982, 0.1213] & 0.9972 & 0.8874 & 0.1520 \\
GNB & Encoded & 0.5163 & [0.5034, 0.5293] & 0.9457 & 0.4294 & 0.4348 \\
RNN & Raw & 0.4667 & [0.4505, 0.4825] & 0.9038 & 0.4371 & 0.4117 \\
RNN & Encoded & 0.7389 & [0.7248, 0.7527] & 0.8789 & 0.1401 & \textbf{0.8764} \\
\midrule
SNN & Encoded & \textbf{0.9873} & \textbf{[0.9856, 0.9888]} &
\textbf{0.9986} & \textbf{0.0113} & 0.8750 \\
\bottomrule
\end{tabular}
\end{table*}

\subsection{Symbolic Feature-Group Ablation}
To assess the relative importance of each symbolic attribute, we performed a structured ablation study during network training. For each dataset, we evolved 200 SNNs: 25 networks for each of the 7 conditions where a single symbolic feature group was removed, and 25 control networks trained with all features. All ablations preserve native inter-event timing during SNN propagation because elapsed time governs recurrent spiking dynamics and is part of the asynchronous computation model. The ablation, therefore, evaluates the contribution of each symbolic input group while keeping the event-time substrate fixed. 

\begin{figure}[t]
  \centering
  \includegraphics[width=.75\columnwidth]{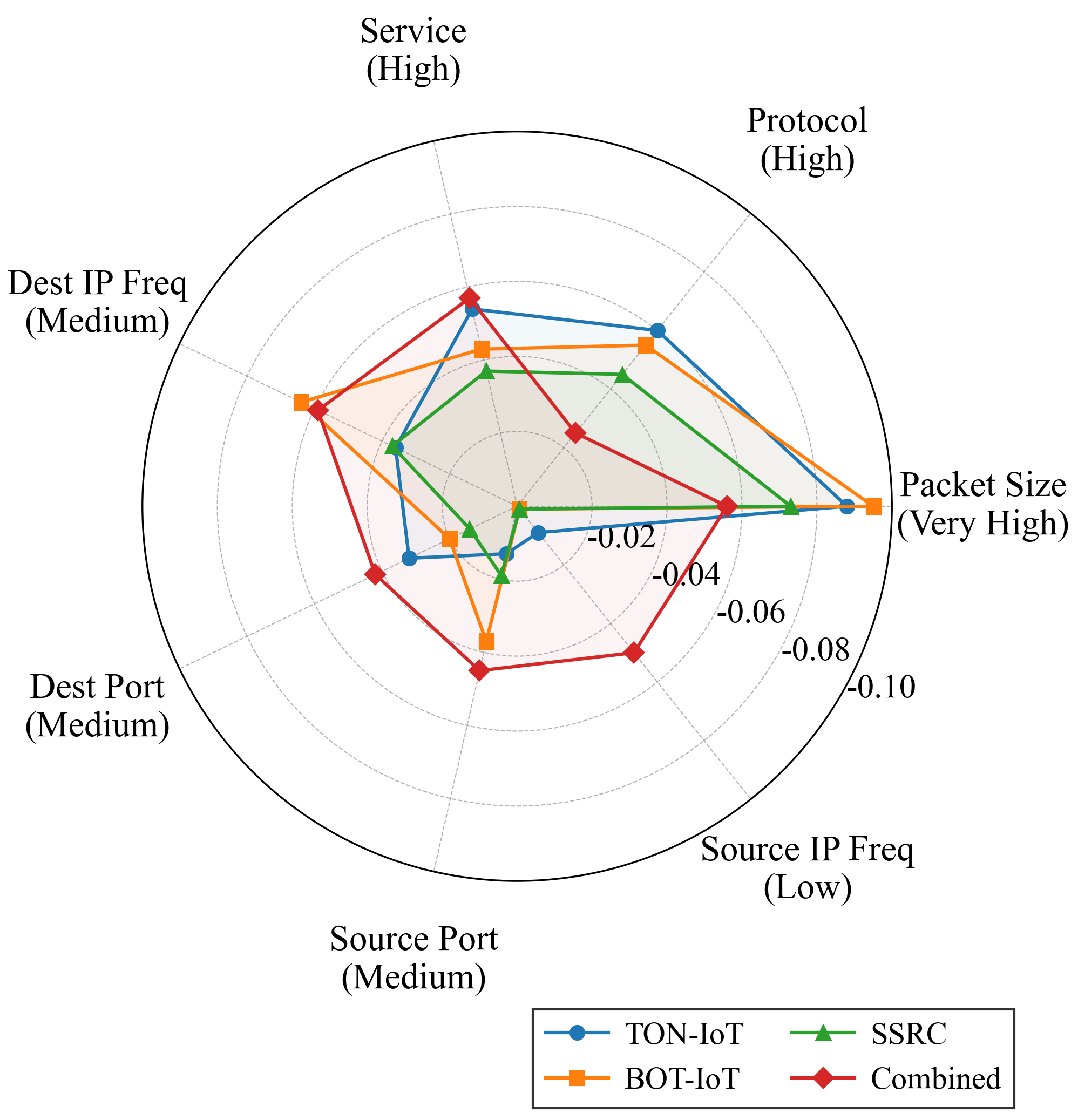}
  \caption{Loss in $J_{hybrid}$ when individual symbolic feature groups are ablated, shown per dataset.}
  \label{fig:ablation}
\end{figure}

As shown in Fig.~\ref{fig:ablation}, packet size and protocol are the most critical symbolic features across all datasets. Ablating packet size leads to the largest degradation (mean $\Delta J_{hybrid} = -0.085$), followed closely by protocol ($\Delta J_{hybrid} = -0.047$). This matches the high volume of small, identical-protocol packet bursts characteristic of DoS and DDoS attacks. Conversely, IP frequency and port groups show negligible impact, suggesting that the networks learn general traffic behaviors rather than overfitting to specific addresses or ports.

\section{CAN IDS Evaluation}
\label{sec:can_ids_evaluation}

To test the transferability of our framework, we apply it to Controller Area Network (CAN) bus traffic, a structurally different asynchronous stream characterized by 11-bit arbitration IDs, short 8-byte payloads, and sub-millisecond inter-message spacing. CAN attacks typically inject spoofed frames to manipulate in-vehicle systems, demanding that a detector analyze both transmission timing and message payload sequences.

\subsection{Dataset and Preprocessing}
We evaluate using two public CAN benchmarks: ROAD~\cite{verma2024road} and Car-Hacking~\cite{song2020carhacking}, alongside a Combined CAN dataset containing runs from both (Table~\ref{tab:can-datasets}). ROAD provides dynamometer captures with long contiguous attack segments, while Car-Hacking represents short, burst-like injection campaigns. For evaluation, adjacent attack messages separated by fewer than 50 benign frames are merged into single attack intervals, and runs are padded with benign context. Splits are 60/10/30 for train/validation/test.

\begin{table}[htbp]
\caption{CAN IDS Dataset Summary}
\label{tab:can-datasets}
\centering
\small
\begin{tabular}{lrrr}
\toprule
Dataset & Runs & Messages & Attack \% \\
\midrule
ROAD & 28 & 25.8 M & 3.06 \\
Car-Hacking & 2,404 & 24.9 M & 9.35 \\
\midrule
Combined CAN & 2,432 & 50.8 M & 6.15 \\
\bottomrule
\end{tabular}
\end{table}

\subsection{Encoding Scheme}
We adapt the three representational roles to CAN semantics, resulting in a 26-dimensional hybrid input vector (Table~\ref{tab:can-encoding}).

\begin{table}[htbp]
\caption{CAN Input Encoding and Input Dimensions}
\label{tab:can-encoding}
\centering
\small
\begin{tabular}{lccp{3cm}}
\toprule
Feature Group & Features & SNN In & Description \\
\midrule
CAN ID freq & 6 & 6 & Sliding 1k window; one-hot: 
0, 1--5, 6--10, 11--20, 21--40, 41+ \\
Payload bytes & 8 & 16 & 8 bytes $\times$ 2 
(flip-flop rate coding) \\
$\Delta t$ & 4 & 4 & ${<}0.5$ ms, $0.5$--$1$ ms, 
$1$--$5$ ms, ${\geq}5$ ms \\
\midrule
\textbf{Total} & \textbf{18} & \textbf{26} & \\
\bottomrule
\end{tabular}
\end{table}

The framework does not require every field to be symbolic. It assigns encoding roles according to field semantics: symbolic identifiers are represented categorically, timing is discretized as $\Delta t$, and payload bytes retain magnitude because their values carry meaningful signal content. This preserves native per-event structure while applying numeric spike encoding only where numeric magnitude is semantically meaningful.

For CAN ID frequency, we count occurrences of each arbitration ID within a sliding 1,000-message window, binning the result into six categorical bins. Relative timing ($\Delta t$) is binned into four classes representing sub-millisecond ranges. For numeric payload content, we employ flip-flop rate coding, assigning two input channels per payload byte to represent its continuous value as positive and negative deviation rates.

\subsection{Results and Baseline Comparison}

The CAN IDS evaluation serves the same purpose as the Network IDS evaluation: to test whether the proposed representation exposes useful structure from native event streams and whether recurrent spiking state adds temporal decision capacity beyond the encoded inputs alone. Unlike packet-level network traffic, CAN messages are shorter, more periodic, and dominated by arbitration ID, payload content, and sub-millisecond inter-message timing. This makes CAN a useful transfer test for the symbolic-temporal framework.

Table~\ref{tab:can-snn} reports the best evolved SNN for each CAN dataset. For each dataset, the reported network was selected as the highest-performing candidate on the training data and then evaluated on held-out test runs. The SNN achieves strong detection on both individual datasets and maintains high performance on the Combined CAN dataset, which merges short injection bursts with longer ROAD captures.

\begin{table}[htbp]
\caption{Best SNN Performance on CAN IDS Datasets}
\label{tab:can-snn}
\centering
\small
\setlength{\tabcolsep}{3pt}
\begin{tabular}{@{}lcccc@{}}
\toprule
Dataset & $J_{hybrid}$ & ADR & $FPR_{event}$ & Message Accuracy \\
\midrule
Car-Hacking & 0.9971 & 1.0000 & 0.0029 & 0.8418 \\
ROAD        & 0.9850 & 1.0000 & 0.0150 & 0.9850 \\
Combined    & 0.9803 & 1.0000 & 0.0197 & 0.9063 \\
\bottomrule
\end{tabular}
\end{table}

These results are not presented as a direct state-of-the-art comparison against all CAN IDS methods, since prior systems often use different preprocessing, feature construction, and evaluation procedures. Existing CAN detectors commonly report accuracies above 99\%~\cite{islam2024unsupervised, rajapaksha2023cansurvey}; the distinction here is that the proposed approach performs detection directly from message arrivals using an edge-aligned SNN and the event-native encoded representation. The lower message-level accuracy on Car-Hacking should be interpreted in light of our operational objective. ADR counts an attack as detected when at least one alarm occurs within its labeled interval, so high ADR can coexist with lower message-level accuracy. We use $J_{hybrid}$ as our primary operational metric for comparing attack detection and benign alarm activity across ROAD, Car-Hacking, and the Combined CAN setting, while reporting message-level accuracy as a complementary measure.

Table~\ref{tab:can-baselines} isolates the contribution of the representation on the Combined CAN dataset. Classical models trained on raw normalized features achieve limited performance. When trained on the proposed 26-dimensional hybrid encoding, performance improves for most baselines. This suggests that the encoding exposes useful structure from CAN ID frequency, payload behavior, and inter-message timing.

\begin{table*}[htbp]
\caption{Baseline Comparison on the Combined CAN Dataset}
\label{tab:can-baselines}
\centering
\small
\begin{tabular}{llccccc}
\toprule
Model & Representation & $J_{hybrid}$ & $J_{hybrid}$ 95\% CI & ADR & $FPR_{event}$ & Message Accuracy \\
\midrule
LR  & Raw     & 0.3741 & [0.3042, 0.4555] & 0.6263 & 0.2522 & 0.7412 \\
LR  & Encoded & 0.5140 & [0.4196, 0.6235] & 0.6824 & 0.1683 & 0.8236 \\
RF  & Raw     & 0.8107 & [0.7221, 0.9183] & 0.8772 & 0.0665 & 0.9304 \\
RF  & Encoded & 0.8410 & [0.7655, 0.9262] & 0.8965 & 0.0555 & 0.9419 \\
DT  & Raw     & 0.8174 & [0.7467, 0.9004] & 0.9115 & 0.0940 & 0.9063 \\
DT  & Encoded & 0.8485 & [0.7910, 0.9167] & 0.9185 & 0.0700 & 0.9294 \\
GNB & Raw     & 0.3049 & [0.2374, 0.3822] & 0.3424 & 0.0375 & 0.9288 \\
GNB & Encoded & 0.2279 & [0.1811, 0.2878] & 0.2516 & 0.0237 & 0.9370 \\
RNN & Raw & 0.9712 & [0.9601, 0.9829] & 0.9918 & 0.0206 & 0.6341 \\
RNN & Encoded & 0.9719 & [0.9579, 0.9830] & 0.9726 & \textbf{0.0007} & \textbf{0.9660} \\
\midrule
SNN & Encoded & \textbf{0.9803} & \textbf{[0.9672, 0.9878]} &
\textbf{1.0000} & 0.0197 & 0.9063 \\
\bottomrule
\end{tabular}
\end{table*}

As in the Network IDS evaluation, the encoded representation improves the performance of static classifiers. A non-spiking RNN evaluated on the same encoded message sequences achieves a $J_{hybrid}=0.9719$ (95\% CI: $[0.9579,\,0.9830]$), compared with $J_{hybrid}=0.9803$ (95\% CI: $[0.9672,\,0.9878]$) for the SNN. These results indicate that temporal recurrence accounts for much of the improvement, while the SNN provides slightly improved performance in an event-driven, hardware-compatible form.

Detection latency is measured as the number of CAN messages between attack onset and the first alarm within the labeled attack interval. On the Combined CAN dataset, the best individual network achieves a median detection latency of 48 messages, with 25\% of attacks detected within 35 messages. At typical CAN bus message rates, this corresponds to short detection delays suitable for real-time edge monitoring.

These results reproduce the central trend observed in the Network IDS study: when native event semantics are preserved in a meaningful representation, SNNs can use temporal state effectively while retaining the low-latency, event-driven computation needed for edge-oriented cyber monitoring.

\section{Discussion}
\subsection{Temporal Validation of the Hybrid Objective}

A limitation of $J_{hybrid}$ is that ADR is interval-based. Long attack intervals provide more opportunities for an alarm to overlap an attack by chance, allowing a temporally uninformative detector to achieve an inflated ADR. We therefore evaluate whether the observed performance reflects meaningful temporal alignment between SNN activity and attack behavior.

We apply a circular-shift control within each held-out run, randomly shifting the SNN output sequence to break its alignment with the attack labels while preserving its activity pattern. The rolling decision rule is then reapplied, and $J_{hybrid}$, ADR, and $FPR_{event}$ are recomputed over 10,000 shifts.

\begin{table}[htbp]
\caption{Observed and Temporally Shifted SNN Performance}
\label{tab:temporal-shift}
\centering
\small
\begin{tabular}{llccc}
\toprule
Application & Evaluation & $J_{hybrid}$ & ADR & $FPR_{event}$ \\
\midrule
Network IDS & Observed
& \textbf{0.9873} & \textbf{0.9986} & \textbf{0.0113} \\
& Shifted Mean
& 0.7540 & 0.8609 & 0.1069 \\
\midrule
CAN IDS & Observed
& \textbf{0.9803} & \textbf{1.0000} & \textbf{0.0197} \\
& Shifted Mean
& 0.6490 & 0.7714 & 0.1224 \\
\bottomrule
\end{tabular}
\end{table}

Temporal shifting reduces mean $J_{hybrid}$ by 0.2333 for Network IDS and 0.3313 for CAN IDS (Table~\ref{tab:temporal-shift}), with the observed score exceeding all 10,000 shifted replicates in both domains ($p<0.0001$). However, shifted ADR remains relatively high, confirming that long attack intervals can still produce chance overlap. We therefore evaluate whether alarms occur promptly and remain aligned with attack activity after onset. First, fixed-horizon analysis measures the cumulative fraction of attacks for which a new alarm episode begins within the first $K$ events. For Network IDS, this reaches 40.5\%, 75.6\%, and 83.1\% within 50, 100, and 250 events, compared with 25.7\%, 31.7\%, and 48.7\% under temporal shifting. For CAN IDS, the corresponding observed rates are 52.1\%, 74.8\%, and 90.1\%, compared with 3.9\%, 7.0\%, and 13.7\%. Median detection latency is also substantially lower for the observed outputs: 55 versus 263 events for Network IDS and 48 versus $>500$ messages for CAN IDS. We additionally examine the instantaneous alarm state at specific positions relative to attack onset. This analysis shows whether alarm activity becomes concentrated after an attack begins. Network IDS increases from 11.0\% of attacks in the alarm state at onset to 31.5\% at event $+49$, while the shifted control remains near 16\%. CAN IDS increases from 0.0\% at onset to 52.1\% at $+49$, while the shifted control remains near 37\%. These results show that the high $J_{hybrid}$ scores are not explained by long attack intervals alone. Observed outputs produce alarms substantially earlier than shifted controls, while new alarm episodes and alarm-state activity are strongly aligned with attack onset.

\subsection{Topology}
\label{sec:discussion}

\begin{figure}[t]
  \centering
  \includegraphics[width=0.8\columnwidth]{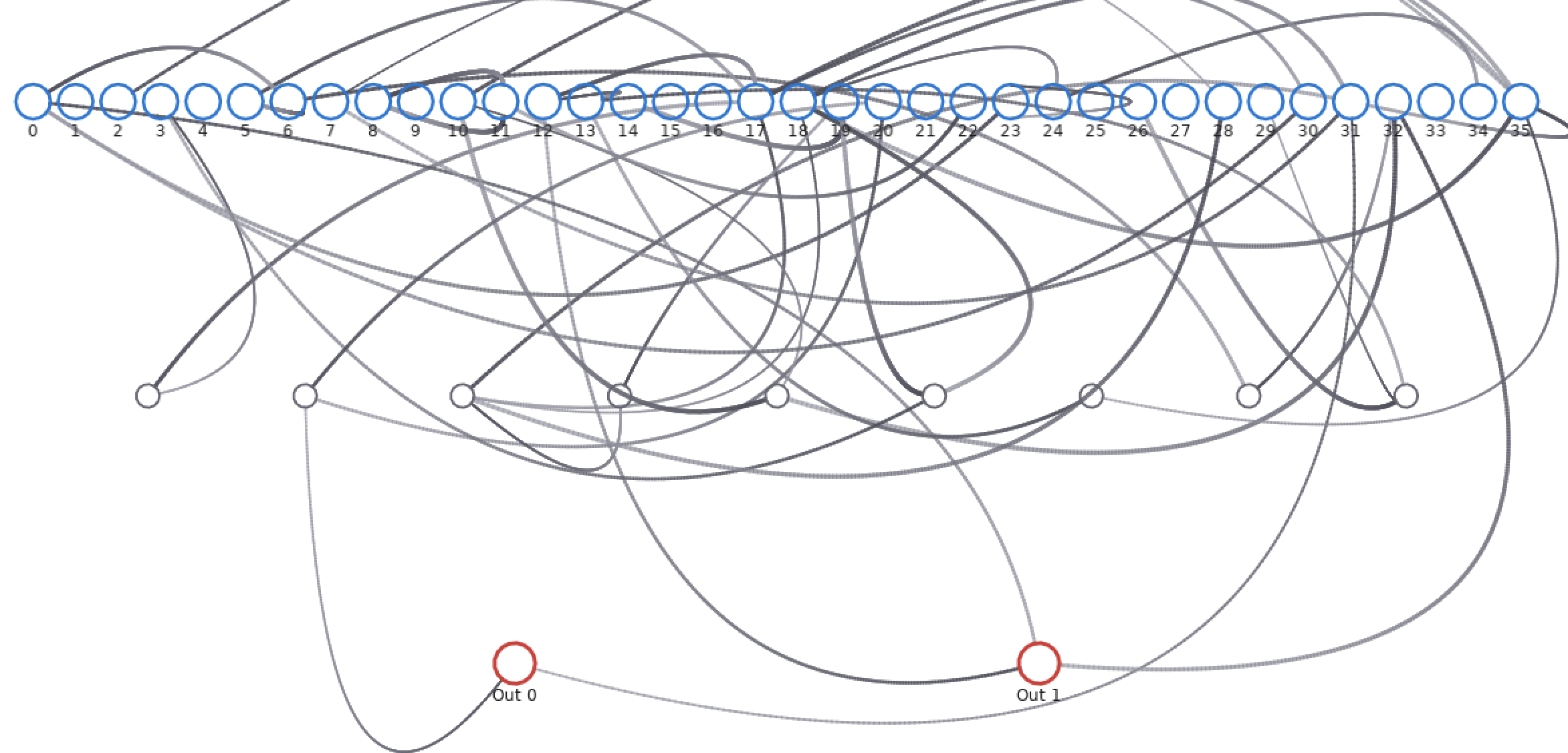}
  \hfill
  \includegraphics[width=0.8\columnwidth]{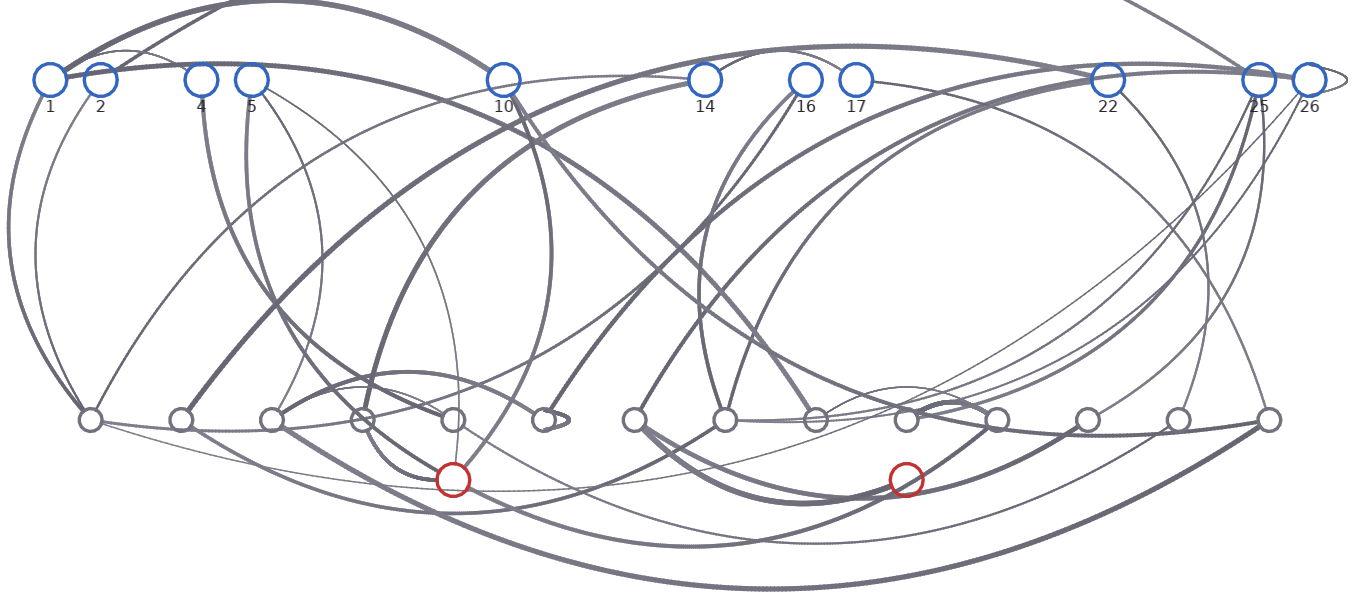}
  \caption{Evolved SNN topologies for Network IDS (top) and CAN IDS (bottom). Input neurons are blue, hidden are grey, and output are red.}
  \label{fig:topologies}
\end{figure}

The evolved SNN topologies also highlight the deployment motivation for the proposed representation. The evolved Network IDS SNN (Fig.~\ref{fig:topologies}, top) uses only 46 neurons with dense recurrence, while still achieving high detection on the Combined dataset. This compact scale contrasts with typical deep IDS architectures, which rely on dense tensor operations and larger parameterized models. The input connections are concentrated around packet size (bins 28/30) and protocol (bins 20/21), which form recurrent hubs. This structural alignment with our ablation study suggests that the network routes information through the most discriminative encoding channels while suppressing brittle categorical indicators.

Similarly, for the CAN IDS network (Fig.~\ref{fig:topologies}, bottom), EONS produced a compact recurrent core using active inputs from ID-frequency bins, payload byte channels, and inter-arrival timing bins. Several inputs are pruned, leaving a small set of connections associated with high-frequency ID transmission and payload deviations. This compact recurrent structure is particularly relevant for CAN monitoring, where short inter-message intervals make detection latency critical: the small topology and event-driven dynamics allow temporal evidence to be processed as messages arrive, without waiting for large windows or dense feature construction.

The compact evolved topologies align with the $\mu$Caspian-oriented deployment constraints~\cite{mitchell2020ucaspian}. $\mu$Caspian is a small event-driven FPGA platform for neuromorphic edge applications. Hardware power and latency have also been evaluated on this platform in prior work: Ghawaly et al.~\cite{Ghawaly_2025} deployed compact EONS-evolved Integrate-and-Fire SNNs on $\mu$Caspian and reported 2~mW power consumption with 20.2~ms inference latency. The resulting Network IDS and CAN IDS topologies fall within the same neuron, synapse, precision, and axonal-delay limits, indicating that the proposed event-native representation supports compact recurrent SNNs compatible with an established edge-oriented neuromorphic deployment path. Although the present work does not report physical hardware measurements for the cyber models, these constraints provide a hardware-aligned basis for future direct deployment from native packet and CAN event streams.

\section{Limitations}
\label{sec:limitations}

We acknowledge several limitations of this study. First, we do not claim that event-native spiking classifiers universally outperform conventional intrusion detection systems. Rather, our contribution is representational: we demonstrate that neuromorphic systems can perform event-native monitoring when cyber semantics are preserved at the input interface. Second, the encoding framework is domain-informed and relies on manually defined bin boundaries. Future work should investigate methods for learning these boundaries directly from data. Third, the RNN and conventional classifiers were tuned through limited sweeps over key hyperparameters rather than exhaustive optimization. Future evaluations could expand this comparison through more extensive tuning and the inclusion of GRUs, LSTMs, and temporal transformers. Additionally, $J_{hybrid}$ is an operational optimization objective rather than a conventional sample-level classification metric; because ADR is interval-based, we supplement it with additional temporal analyses to test whether detections are genuinely aligned with attack activity. Finally, our evaluations use public benchmark datasets and simulated hardware constraints; therefore, they do not capture all deployment challenges.

\section{Conclusion}
\label{sec:conclusion}

This paper addresses the representational gap at the intersection of spiking neural networks and cybersecurity. We introduced a symbolic-temporal spike encoding framework that maps heterogeneous, asynchronous cyber events directly to sparse, spike-compatible vectors at arrival. By assigning distinct encoding roles to semantic identity, local frequency context, and inter-event timing, the framework preserves categorical semantics and temporal spacing without converting events to artificial numeric averages. 

Across packet-level Network IDS and message-level CAN IDS, compact recurrent SNNs operating under edge-aligned $\mu$Caspian constraints achieved strong detection ($J_{hybrid} \geq 0.980$). The proposed event-to-spike representation interface preserves the symbolic and temporal structure of cyber streams, enabling edge-aligned SNNs to perform direct, low-latency neuromorphic monitoring on native event streams.

\bibliographystyle{IEEEtran}
\bibliography{references}

\end{document}